\documentclass[11pt]{article} 
\usepackage{times}
\usepackage{psfig}

\RequirePackage{ifthen}
\RequirePackage{times}
\RequirePackage{mathptm}
\RequirePackage{pifont}

\usepackage{fullpage}

\begin{document}

\title{Differential Methods in Catadioptric Sensor Design with
Applications to Panoramic Imaging\\
{\small Technical Report}
}

\author{R. Andrew Hicks\\
Department of Mathematics\\ 
Drexel University \\ 
ahicks@drexel.edu\\
}

\maketitle

\begin{abstract}
We discuss design techniques for catadioptric sensors that realize
given projections. In general, these problems do not have solutions,
but approximate solutions may often be found that are visually
acceptable. There are several methods to approach this problem, but
here we focus on what we call the ``vector field approach''. An
application is given where a true panoramic mirror is derived, i.e. a
mirror that yields a cylindrical projection to the viewer without any
digital unwarping.
\end{abstract}

\thispagestyle{empty}

\section{Introduction}

A fundamental problem of imaging is the creation of wide-angle and
panoramic images. Wide-angle images that do not have radial distortion
require complex assemblies of lenses, because as the field of view
increases, abberations are introduced, which in require the addition
of more lenses for correction. Fish-eye lenses offer wide-field of
views but at the cost of considerable radial distortion. This
distortion may be removed in software, but the resulting images suffer
from poor resolution in some regions.

Panoramic images present a related challenge. A straightforward, yet
tedious approach is the stitching together of images. In the 19th
century numerous cameras were created for taking panoramic still shots
through ingenious rotating mechanisms (e.g. rotating slit cameras),
but were inherently awkward. Adapting such devices for digital
imaging (especially video applications) is clearly problematic.

The introduction of a curved mirror into a conventional imaging system
is an elegant solution to these problems. General design questions are: 
\smallskip

{ 
\noindent
A. How does one design a catadioptric sensor to image a prescribed
region of space ?
\smallskip

\noindent
B. How does one design a catadioptric sensor to maximize image quality ?
\smallskip

\noindent 
C. How does one design an optimal catadioptric sensor for a
given machine vision application, such as robot navigation, tracking
or segmentation ?  
\smallskip

Commercial interest in panoramic sensors based on these ideas has
risen sharply in the last five years, and a large number of companies
have emerged to to meet the demand. Common commercial applications are
surveillance and ``virtual tours'' on the web. At last count, at least
19 different companies were found offering panoramic sensors for sale,
the majority of which were based on catadioptrics (see
\cite{omnipage}).

\section{Related Work}

\subsection{Rotationally Symmetric Mirrors}

Almost all research performed with mirrors refers to {\bf rotationally
symmetric mirrors}, since these mirrors are the simplest to make and
to mathematically model. For every curve in the plane, one may create
a mirror by rotating the curve about a chosen line, which will then
play the role of the optical axis. Clearly then, there are an infinite
number of possible mirrors and the question then is which one to
choose for a given application. This generally gives rise to the
problem of finding a mirror shape given a prescribed property. To do
this generally amounts to solving a differential equation. For
example, if one requires a curve that focuses parallel rays, the
answer is a parabola. In this case, the curve and it's special
property are familiar and differential methods are not needed, but
this is a rare exception.

An early use of mirrors for panoramic imaging is a patent by Rees
\cite{rees70}, who proposes the use of a hyperbolic mirror for
television viewing. Another patent is by Greguss \cite{greguss86},
which is a system for panoramic viewing based on an annular lens
combined with mirrored surfaces. 

Early applications to robotics using a conical mirror were carried by
Yagi et al. in \cite{yagi90}.  Yamazawa et al \cite{yamazawa93} use a
hyperbolic mirror for obstacle detection.  

In \cite{nayar97cvpr}, Nayar describes an omnidirectional sensor from
which it is possible to create perspective views by unwarping the
images in software. In order to do this, the sensor must satisfy the
{\bf single viewpoint constraint}, which means that the sensor only
measures the intensity of light passing through some fixed single
point in space. This sensor uses a parabolic mirror, which,
remarkably, is the only shape from which one can achieve a perspective
unwarping of the image when using an orthographic camera (see
\cite{baker98iccv}).

The first use of differential methods for design was by Chahl and
Srinivasan in \cite{chahl97}. Here the authors employ differential
methods to derive a mirror shape such that the radial angle $\theta$
and the angle of elevation $\phi$ (see figure (\ref{fig:chahlmirror}))
are linearly related.

\begin{figure}[ht]
\centerline{
\psfig{figure=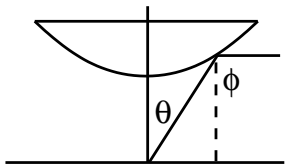,width=1.5in}}
\caption{Chahl and Srinivasan consider a mirror in which
$\frac{d\phi}{d\theta}$ is a constant $K$.}
\label{fig:chahlmirror}
\end{figure}

In \cite{ollis99}, Ollis et al. argue that improved resolution
uniformity is obtained if one demands that the angle of elevation
$\phi$ be proportional to the {\it tangent} of the radial angle
$\theta$; geometrically, $\tan(\theta)$ represents the distance of the
image point to the optical axis. Thus, as the angle $\theta$ "sweeps"
out some range of angle, a proportional length is swept out in the
image plane (note that the line of reasoning here is essentially one
dimensional). The authors then apply these mirrors to
mobile robot navigation and panoramic stereo.

A variant of the Ollis mirror is described by Conroy and Moore in
\cite{conroy99iccv}. Using the same parameters as above, the authors
then calculate what the mirror shape such that $\phi$ is proportional
to the area of the corresponding disk in the image plane.

Hicks and Bajcsy derive the mirror shapes for a sensor in
\cite{hicks00cvpr} which will give wide-angle perspective images, and
so uniformly image planes. In \cite{hicks02omnivis} Hicks and Perline
describe a sensor for which the projection map from the
view sphere to the image plane is area preserving. This means that any
two solid angles are allocated the same number of pixels by the
sensor. 

Sensors designed to image cylinders were investigated by Gaechter and
Pajdla in \cite{gaechter01icar}. The analysis of these devices is
essentially two dimensional and extended to three dimensions by radial
symmetry. In fact, the unwarping map (corresponding to a cylindrical
projection) for these sensors is the standard polar map.

\subsection{The Asymmetric Case}

If one leaves the realm of rotational symmetry, the problem of
constructing a mirror for a given projection gets significantly more
difficult. Not only are the equations much more complex, but with
probability 1 they will not have solutions.

In \cite{hicks01cvpr} this problem is addressed and methods for
approximating solutions are introduced. Here we extend this work and
give a new application.

\section{Problem Statement} 

We work under the assumption that all cameras realize perfect
perspective or orthographic projections.
\smallskip

Suppose one is given a fixed surface $S$ in $R^3$ and a camera with
image plane $I$, also fixed in $R^3$. A given mirrored surface $M$
induces a transformation $T_M$ from some subset of $I$ to $S$ by
following a ray from a point ${\bf q} \in I$ until it intersects the
mirror at a point ${\bf r}$. The ray is then reflected according to
the usual law that the angle of incidence is equal the angle of
reflection and intersects $S$ at a point ${\bf s}$. We then define
$T_M({\bf q}) = {\bf s}$ .
\smallskip

\begin{figure}[ht]
\centerline{
\psfig{figure=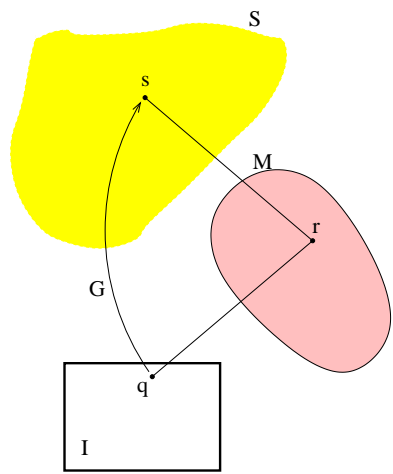,width=2in}}
\caption{The statement of the most general form of the problem: given
a transformation $G$ from an image plane $I$ to a surface $S$, find a
mirror such that the induced optical transformation is as close to the
prescribed transformation $G$ as possible.}
\label{fig:general-problem}
\end{figure}

\noindent
The general problem is:
\medskip
 
\noindent
\fbox{
\parbox{6.3in}{Given $G:I \longrightarrow S$, find $M$ such that
$T_M=G$. If no such $M$ exists, then find $M$ such that $T_M$ is a good
approximation to $G$.}  }
\medskip

\noindent
Figure (\ref{fig:general-problem}) is a corresponding diagram. If an
exact solution exists, then there are several ways to calculate
it. Otherwise, there are numerous ways to formulate and solve the
approximation problem.  Below we briefly discuss three of these
approaches - the fixed surface method, the vector field method, and
the method of distributions . If the problem does not have a solution,
then each formulation gives rise to many possible choices of
approximation techniques.

\subsection{The Fixed-Surface Method}

In this method, one assumes that a generic $M$ is given in some
coordinates and then from it calculate $T_M$. If one represents $M$ as
a graph $z=f(x,y)$ then the expression for $T_M$ will contain partial
derivatives of $f$. Thus the problem of finding $M$ such that $T_M =
G$ is reduced to the problem of simultaneously solving a system of
partial differential equations.  This is the approach introduced in
\cite{hicks01cvpr}.

\subsection{The Vector-field Method}

Notice that for a given $M$, with ${\bf q, r}$ and ${\bf s}$ as above,
that the vector $\frac{{\bf q}-{\bf r}} {|{\bf q}-{\bf r}|} +
\frac{{\bf s}-{\bf r}} {|{\bf s}-{\bf r}|} $ is normal to $M$ at ${\bf
r}$. This suggest a method of constructing a vector field ${\bf W}$ on
$R^3$ that will be normal to the solution: for each ${\bf r} \in R^3$
define

\begin{equation}
{\bf W}({\bf r})= \frac{{\bf q}({\bf r})-{\bf r}} {|{\bf q({\bf r})}-{\bf r}|} + \frac{{\bf
G(q({\bf r}))}-{\bf r}} {|{\bf G(q({\bf r}))}-{\bf r}|}
\end{equation}

\noindent
where ${\bf q}$ is the point in the image plane $I$ corresponding to
${\bf r}$ (i.e. rays traced out from ${\bf q}$ contain ${\bf r}$). If
$M$ exists and is expressed as a level surface $F(x,y,z)=0$, then
there will be a scalar function $\lambda(x,y,z)$ s.t.

\begin{equation}
\nabla F = \lambda {\bf W}.
\end{equation}

In fact, several necessary and sufficient condition exist for testing
for the existence of $M$ from ${\bf W}$. For example, $M$ exists iff
$(\nabla \times {\bf W})\cdot{\bf W}=0$.

\subsection{The Method of Distributions}

In the above vector field formulation clearly ${\bf W}$ as defined is
only parallel to $\nabla F$ and not generally equal to it. Thus the
underlying structure of importance is the planar distribution $\cal D$
that is orthogonal to ${\bf W}$. Hence another formulation of the
problem is to find an integral surface of the distribution $\cal D$
(which is, of course, defined in terms of ${\bf W}$). This may be then
phrased as a variational problem, which may be attacked directly with
numerical methods or one may consider the corresponding Euler-Lagrange
equations. We will not discuss this method any further or give any
examples here.

\section{Analog Panoramic Imaging}

The projection we are interested in mimicking with a mirror is the
so-called cylindrical projection, in which the world is first
projected onto a cylinder, which is the cut and rolled open, giving a
panoramic ``strip'' (which for our purposes will lie in $I$). The
resulting strip can be thought of as having coordinates $r$ and
$\theta$ where the $\theta$ coordinate varies from $0$ to $2\pi$ or
some equivalent interval. Of course we cannot realize this projection
exactly, since it has a single projection point, but we can
approximate it very well.

We take as our object surface $S$ a cylinder whose axis of symmetry
coincides with the optical axis of our camera, which we assume
realizes an orthographic projection. This will allow us to apply the
vector field approach. Despite the fact that we will use the vector
field ${\bf W}$ described above, and not attempt to find an optimal
$\lambda$, the result is visually acceptable.

In figure (\ref{fig:birds}) appears three views of a mirror that has
this property. This mirror provides a $\pm 30$ degree view in the
vertical and a full 360 degree view in the horizontal. In figure
(\ref{fig:chess}) we see this surface used to image a chess board. To
achieve this view the mirror is placed in the middle of the board (at
approximately the height of the king's head) and viewed from below.

\begin{figure}[ht]
\centerline{
\psfig{figure=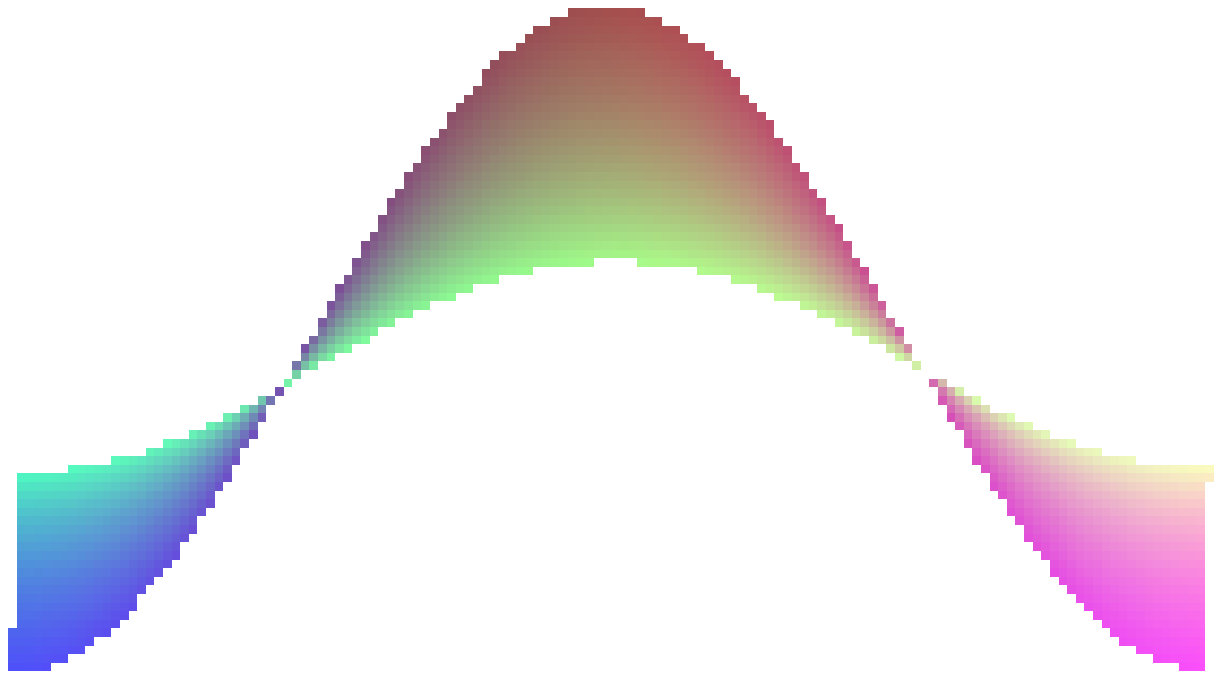,width=1.5in}
\psfig{figure=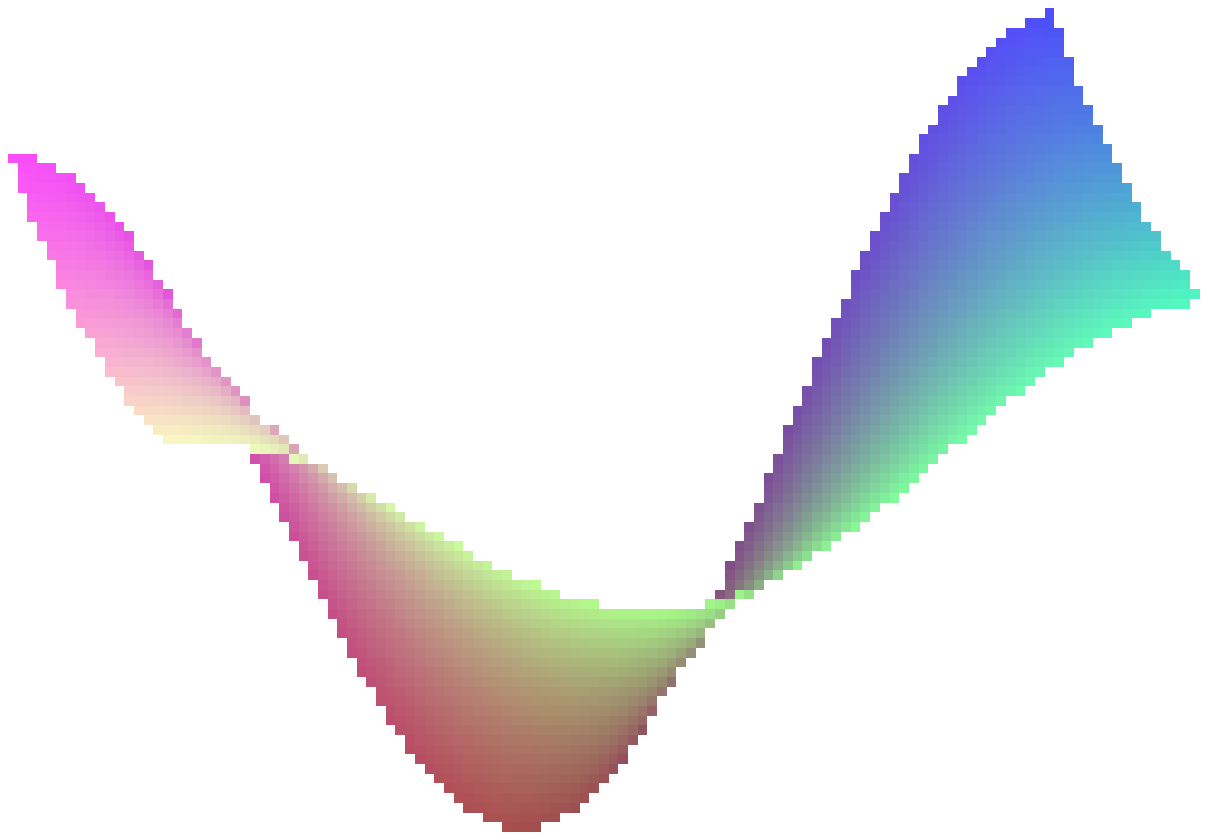,width=1.5in} 
\psfig{figure=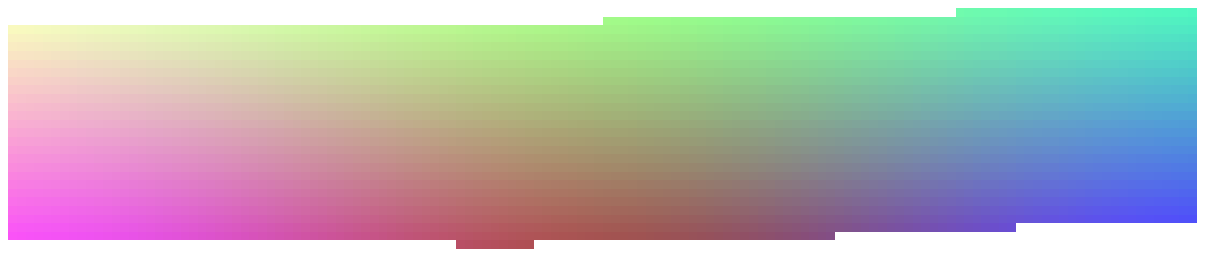,width=1.5in}}
\caption{Three different views of a surface which when used as a
mirror gives a panoramic view without any digital unwarping.}
\label{fig:birds}
\end{figure}

\begin{figure}[ht]
\centerline{
\psfig{figure=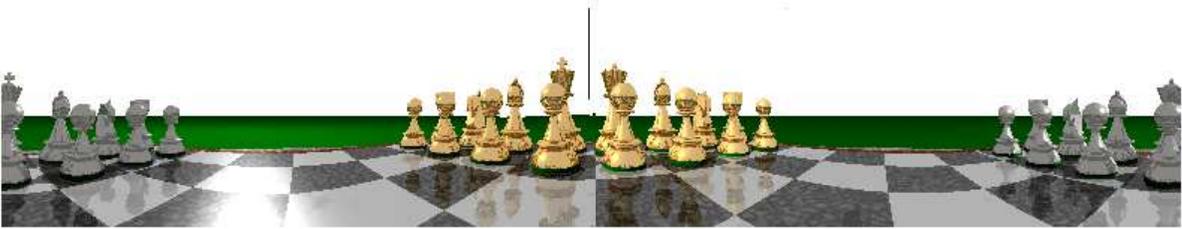,height=1.2in}}
\caption{A panoramic view of a chessboard using the above mirror.}
\label{fig:chess}
\end{figure}

Possible advantages of using a panoramic mirror of the above type over
existing systems are that this sensor would not require a digital
computer for unwarping, and that by performing the unwarping prior to
the sampling of the image, the resolution of the image is more
uniform.

\begin{figure}[ht]
\centerline{
\psfig{figure=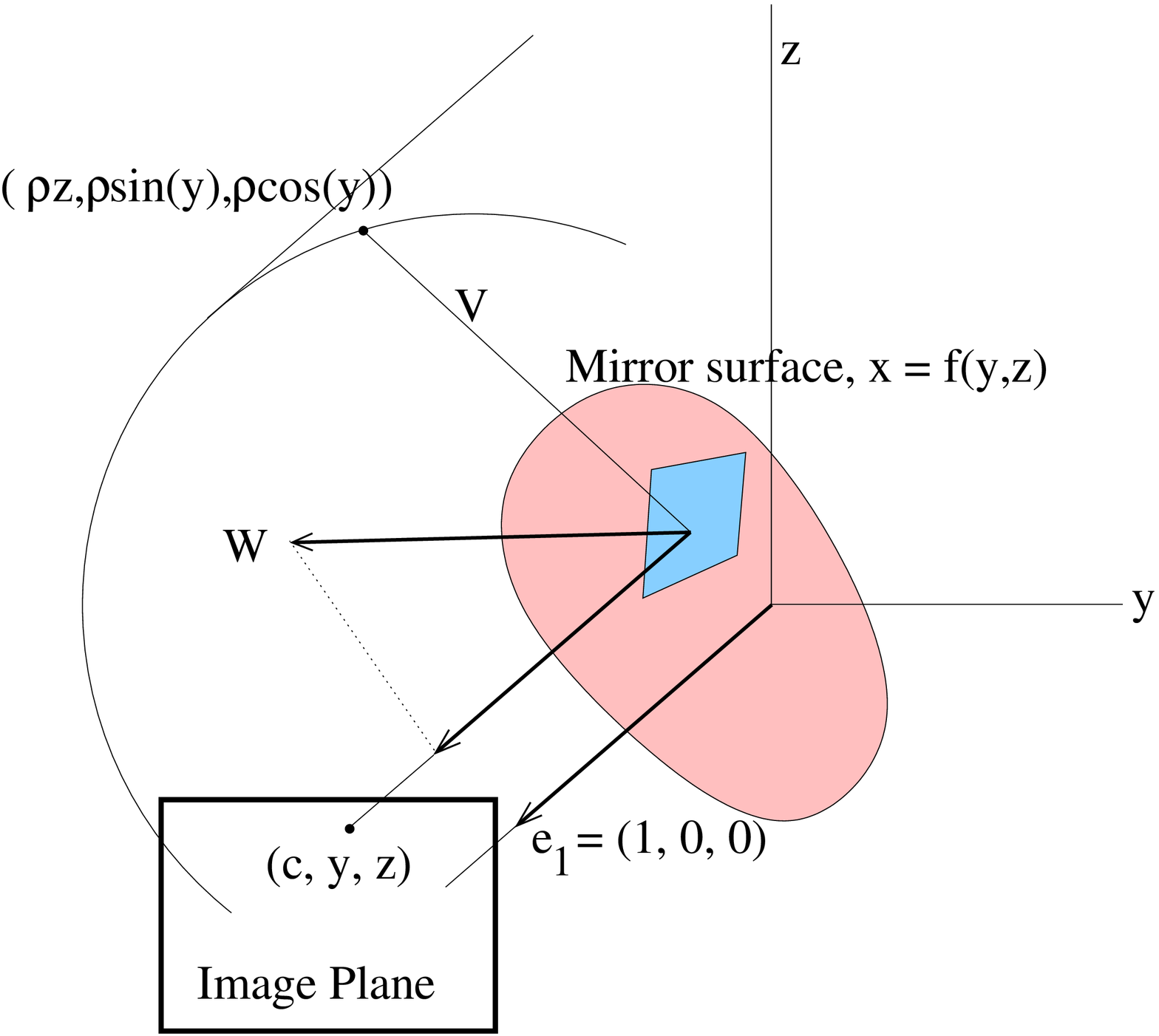,width=3in}}
\caption{}
\label{fig:cylinder}
\end{figure}

To find the above surface, we take the image plane, $I$, to be of the
form $x=c$, so the optical axis is the $x$-axis. The surface $S$ is
taken to be a cylinder of radius $\rho$ about the $x$-axis. The vector
field ${\bf W}$ is derived as follows. According to our above
formulation, ${\bf r} = \left[ x,y, z \right], {\bf q} = \left[ c,y,z
\right]$, and ${\bf s} = \left[ \rho z,\rho\sin(y), \rho \cos(y)
\right]$. ($z$ plays the role of $r$ and $y$ the role of $\theta$). At
${\bf r}$ the incident ray of light should be in the direction ${\bf
V}$ where

\begin{equation}
{\bf V} = \left[ \rho z,\rho\sin(y), \rho \cos(y) \right] - \left[
x,y,z \right].
\end{equation}

\noindent
(See figure (\ref{fig:cylinder}).)   Normalizing ${\bf V}$ and letting $\rho
\rightarrow \infty$ gives the unit vector

\begin{equation} 
{\bf U } = \left[ \frac{z}{\sqrt{1+z^2}}, \frac{\sin(y)}{\sqrt{1+z^2}},
\frac{\cos(y)}{\sqrt{1+z^2}} \right].
\end{equation}

Then ${\bf W} = {\bf U} + {\bf e_1}$, i.e.,

\begin{equation}
{\bf W} = \left[ \frac{z}{\sqrt{1+z^2}} + 1, \frac{\sin(y)}{\sqrt{1+z^2}},
\frac{\cos(y)}{\sqrt{1+z^2}} \right]
\end{equation}

Since we are free to scale ${\bf W}$ as we see fit and $\nabla
(x-f(y,z)) = \left[ -1, f_y, f_z \right]$, a natural choice is to
scale by the inverse of the first component of ${\bf W}$, in which
case the resulting vector field is

\begin{equation}
\left[ 1, \frac{\sin(y)}{z+ \sqrt{1+z^2}}, \frac{\cos(y)}{z+\sqrt{1+z^2}}  
\right].
\end{equation}

\noindent
Thus the problem is to find a function $f(y,z)$ whose gradient is as
close to $\left[ \frac{-\sin(y)}{z+ \sqrt{1+z^2}},
\frac{-\cos(y)}{z+\sqrt{1+z^2}} \right]$ as possible. Clearly there
are different notions of closeness, but a very natural solution is to
take $g= \frac{-\sin(y)}{z+ \sqrt{1+z^2}}$ and $h=
\frac{-`\cos(y)}{z+\sqrt{1+z^2}}$ and then take $f$ to be the
minimizer of

\begin{equation}
\int_A (f_y-g)^2 + (f_z-h)^2 dA.
\label{eqn:integral}
\end{equation}

\noindent
where $A$ is a region in the $y-z$ plane corresponding to the portion
of th image plane $I$ that is of interest (the panoramic
strip)\footnote{This is closely related to the {\bf Hodge
decomposition theorem}, which states that every vector field may be
orthogonally decomposed into an exact vector field $\nabla f$ plus a
vector field with zero divergence. The proof boils down to considering
Poisson's equation with known boundary conditions, and so existence
and uniqueness is easily established. This provides an alternative
computational approach to the problem.}. Presumably though we wish the
$y$ width of $A$ to be at least $2\pi$ in order to get a panoramic
image. The $z$ component will control to vertical field of view. The
above integral was minimized in Maple by taking $f$ was to be a
generic polynomial in $y$ and $z$ of fixed degree.  Notice that the
surface used in the ray tracing examples above was derived taking
$\rho \rightarrow \infty$, and yet it works well for ``close'' objects
- the mirror itself was approximately the size of a chess piece.
\medskip

Finally, the above sensor can be altered to fill an entire 640 by 480
video image entirely. In figure (\ref{fig:helmets}) we see several
images of a mirror constructed by taking the union of the section of
the above mirror corresponding to $-\frac{\pi}{2} \leq y \leq
\frac{\pi}{2}$ with its reflection, resulting a shape reminiscent of a
conquistador helmet. This means that all the pixels of the video
camera are being used to create the panoramic image. Compared with
rotationally symmetric systems these lead to a 70 \% increase in image
resolution. Figure (\ref{fig:conq}) shows the chess board using this
mirror.

\begin{figure}[ht]
\centerline{
\psfig{figure=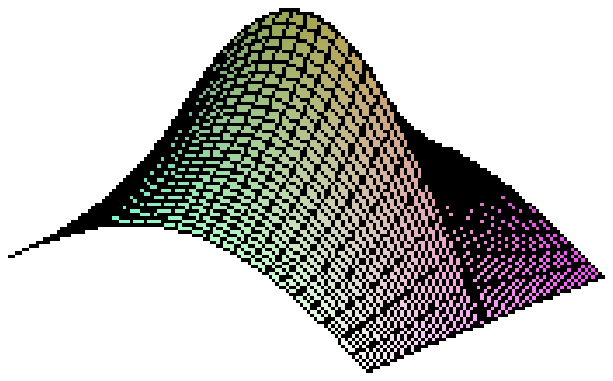,width=2in}
\psfig{figure=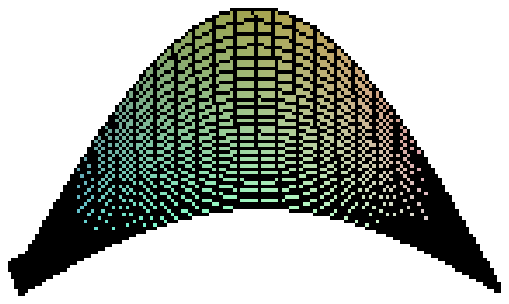,width=2in} 
\psfig{figure=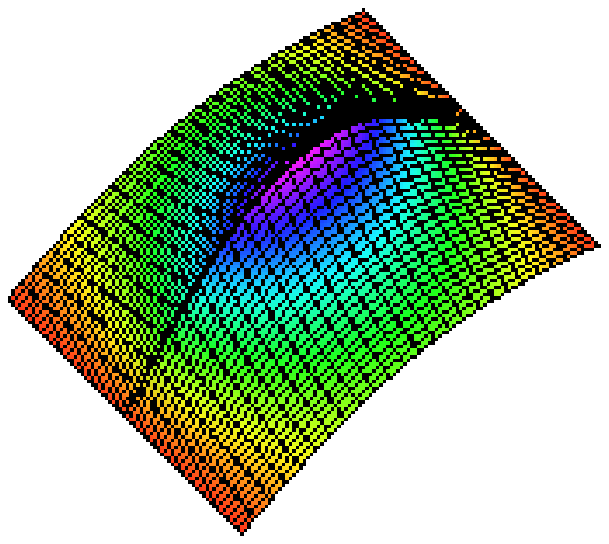,width=2in}}
\caption{A panoramic ``conquistador'' mirror,  designed with proportions close to 3:4.}
\label{fig:helmets}
\end{figure}

\begin{figure}[ht]
\centerline{
\psfig{figure=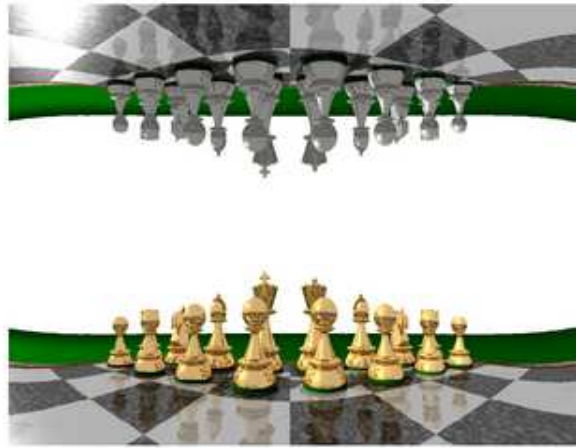,width=3in}
}
\caption{The chessboard viewed with the conquistador mirror.}
\label{fig:conq}
\end{figure}

\renewcommand{\baselinestretch}{1.0} 

\bibliography{compomni} 

\begin{thebibliography}{10}

\bibitem{baker98iccv}
S.~Baker and S.~Nayar.
\newblock A theory of catadioptric image formation.
\newblock In {\em Proc. International Conference on Computer Vision}, pages
  35--42, 1998.

\bibitem{chahl97}
J.S. Chahl and M.V. Srinivasan.
\newblock Reflective surfaces for panoramic imaging.
\newblock {\em Applied Optics}, 36:8275--8285, 1997.

\bibitem{conroy99iccv}
T.~Conroy and J.~Moore.
\newblock Resolution invariant surfaces for panoramic vision systems.
\newblock In {\em Proc. International Conference on Computer Vision}, pages
  392--397, 1999.

\bibitem{omnipage}
K.~Daniilidis.
\newblock The page of omnidirectional vision,
  http://www.cis.upenn.edu/~kostas/omni.html, 2002.

\bibitem{gaechter01icar}
S.~Gaechter and T.~Pajdla.
\newblock Mirror design for an omnidirectional camera with space variant
  imager.
\newblock In {\em Proc. of the Workshop on Omnidirectional Vision Applied to
  Robotic Orientation and Nondestructive Testing (NDT), Budapest}, 2001.

\bibitem{greguss86}
P.~Greguss.
\newblock {\em Panoramic Imaging Block for Three-dimensional space}.
\newblock United States Patent, (4,566,736), January, 1986.

\bibitem{hicks00cvpr}
R.~A. Hicks and R.~Bajcsy.
\newblock Catadioptic sensors that approximate wide-angle perspective
  projections.
\newblock In {\em Proc. Computer Vision Pattern Recognition}, pages 545--551,
  2000.

\bibitem{hicks01cvpr}
R.~A. Hicks and R.~Perline.
\newblock Geometric distributions and catadioptric sensor design.
\newblock In {\em Proc. Computer Vision Pattern Recognition}, pages 584--589,
  2001.

\bibitem{hicks02omnivis}
R.~A. Hicks and R.~Perline.
\newblock Equi-areal catadioptric sensors.
\newblock In {\em Proc. of IEEE Workshop on Omnidirectional Vision}, pages
  13--18, 2002.

\bibitem{nayar97cvpr}
S.~Nayar.
\newblock Catadioptric omnidirectional camera.
\newblock In {\em Proc. Computer Vision Pattern Recognition}, pages 482--488,
  1997.

\bibitem{ollis99}
M.~Ollis, H.~Herman, and Sanjiv Singh.
\newblock Analysis and design of panoramic stereo vision using equi-angular
  pixel cameras.
\newblock {\em Technical Report, The Robotics Institute, Carnegie Mellon
  University, 5000 Forbes Avenue Pittsburgh, PA 15213}, 1999.

\bibitem{rees70}
D.~Rees.
\newblock {\em Panoramic television viewing system}.
\newblock United States Patent, (3,505,465), April, 1970.

\bibitem{yagi90}
Y.~Yagi and S.~Kawato.
\newblock Panoramic scene analysis with conic projection.
\newblock In {\em Proceedings of the International Conference on Robots and
  Systems}, 1990.

\bibitem{yamazawa93}
K.~Yamazawa, Y.~Yagi, and M.~Yachida.
\newblock Omnidirectional imaging with hyperboidal projection.
\newblock In {\em Proceedings of the IEEE International Conference on Robots
  and Systems}, 1993.

\end{thebibliography}
\bibliographystyle{plain} 

\end{document}